\documentclass[12pt]{article}

\usepackage{graphicx}
\usepackage{booktabs}
\usepackage{hyperref}
\usepackage[margin=1in]{geometry}
\usepackage{float}  

\title{Asia Cup 2025: A Structured T20 Match-Level Dataset and Exploratory Analysis for Cricket Analytics}

\author{
\textbf{Kousar Raza$^{1}$}\thanks{ORCID: \href{https://orcid.org/0009-0007-5298-4469}{0009-0007-5298-4469}} \quad
\textbf{Faizan Ali$^{2}$} \\[0.8em]
$^{1}$Department of Applied Mathematics and Computer Science,\\ University of Isfahan, Iran\\
$^{2}$Institute of Mathematics and Computer Science,\\ University of Sindh, Jamshoro, Pakistan
}

\date{}
\renewenvironment{abstract}
  {\begin{center}
   \bfseries Abstract
   \end{center}
   \rmfamily\normalsize}
  {}

\begin{document}

\maketitle

\begin{abstract}
This paper presents a structured and comprehensive dataset corresponding to the 2025 Asia Cup T20 cricket tournament, designed to facilitate data-driven research in sports analytics. The dataset comprises records from all 19 matches of the tournament and includes 61 variables covering team scores, wickets, powerplay statistics, boundary counts, toss decisions, venues, and player-specific highlights. To demonstrate its analytical value, we conduct an exploratory data analysis focusing on team performance indicators, boundary distributions, and scoring patterns. The dataset is publicly released through Zenodo under a CC-BY 4.0 license to support reproducibility and further research in cricket analytics, predictive modeling, and strategic decision-making. This work contributes an open, machine-readable benchmark dataset for advancing cricket analytics research.
\end{abstract}\\



\section{Introduction}
Cricket analytics increasingly relies on high-quality datasets to support statistical analysis and machine learning applications. While ball-by-ball data is available for major tournaments, openly accessible and well-documented match-level datasets for specific international competitions remain limited. The Asia Cup 2025 T20 tournament provides a valuable context for analyzing high-pressure, short-format cricket. This paper introduces the \textbf{Asia Cup 2025 Match-Level Dataset}, developed to support reproducible research in sports analytics.


\section{Dataset Description}
The dataset is distributed as a single CSV file, \texttt{asia\_cup\_2025\_complete\_dataset.csv}, containing records for all 19 matches of the tournament with 61 attributes per match.

\subsection{Data Structure and Variables}
Table~\ref{tab:variables} summarizes the major variable categories included in the dataset.

\begin{table}[h!]
\centering
\begin{tabular}{p{6cm} p{8cm}}
\toprule
\textbf{Category} & \textbf{Representative Variables} \\
\midrule
Match Identification & Match\_Number, Date, Venue, Series \\
Teams & Team1, Team2, Playing\_XI \\
Toss Information & Toss\_Winner, Toss\_Decision \\
Match Outcome & Result, Win\_Margin, Player\_of\_the\_Match \\
Team Performance & Total\_Runs, Wickets, Overs, Extras \\
Powerplay Statistics & Powerplay\_Runs, Powerplay\_Overs \\
Boundary Data & Fours, Sixes \\
Tournament Context & Stage, Group, Match\_Format \\
\bottomrule
\end{tabular}
\caption{Summary of variable categories in the Asia Cup 2025 dataset}
\label{tab:variables}
\end{table}

\subsection{Data Sources and Collection}
Match data were collected from publicly available scorecards on \textbf{ESPNcricinfo} and cross-verified with official tournament information released by the \textbf{Asian Cricket Council (ACC)}. Data processing involved manual verification, standardization of entity names, and consistency checks across sources.


\section{Exploratory Analysis}
Exploratory analysis was conducted to demonstrate the analytical potential of the dataset, focusing on toss outcomes, team batting performance, and boundary distributions.

\subsection{Toss Impact Analysis}

\begin{figure}[H]  
\centering
\includegraphics[width=0.8\textwidth]{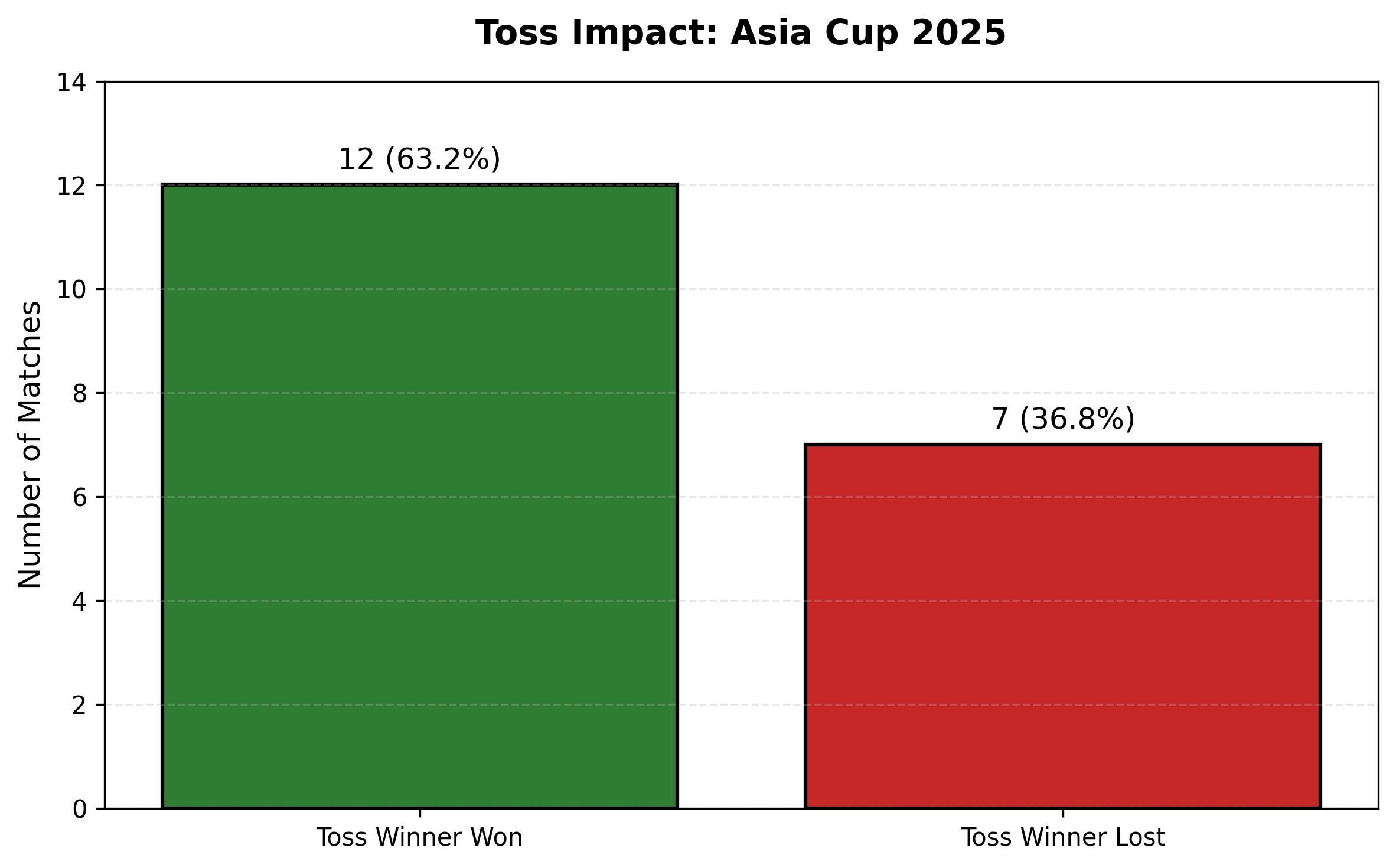}
\caption{Impact of toss outcome on match results in the Asia Cup 2025 tournament.}
\label{fig:toss}
\end{figure}

\subsection{Team Batting Performance}

\begin{figure}[H]
\centering
\includegraphics[width=0.9\textwidth]{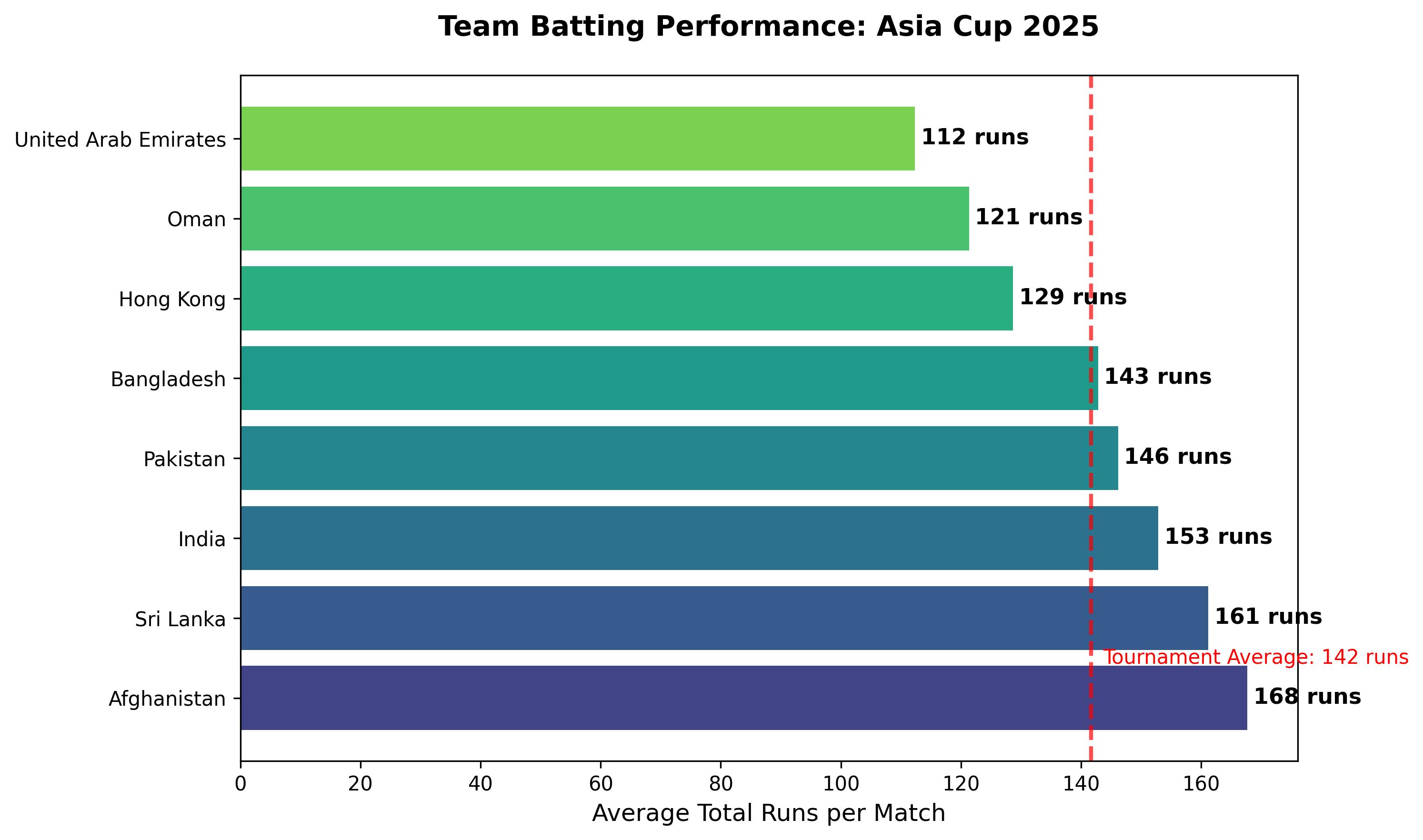}
\caption{Average team scores across all Asia Cup 2025 matches.}
\label{fig:teams}
\end{figure}

\subsection{Boundary Analysis}

\begin{figure}[H]
\centering
\includegraphics[width=0.7\textwidth]{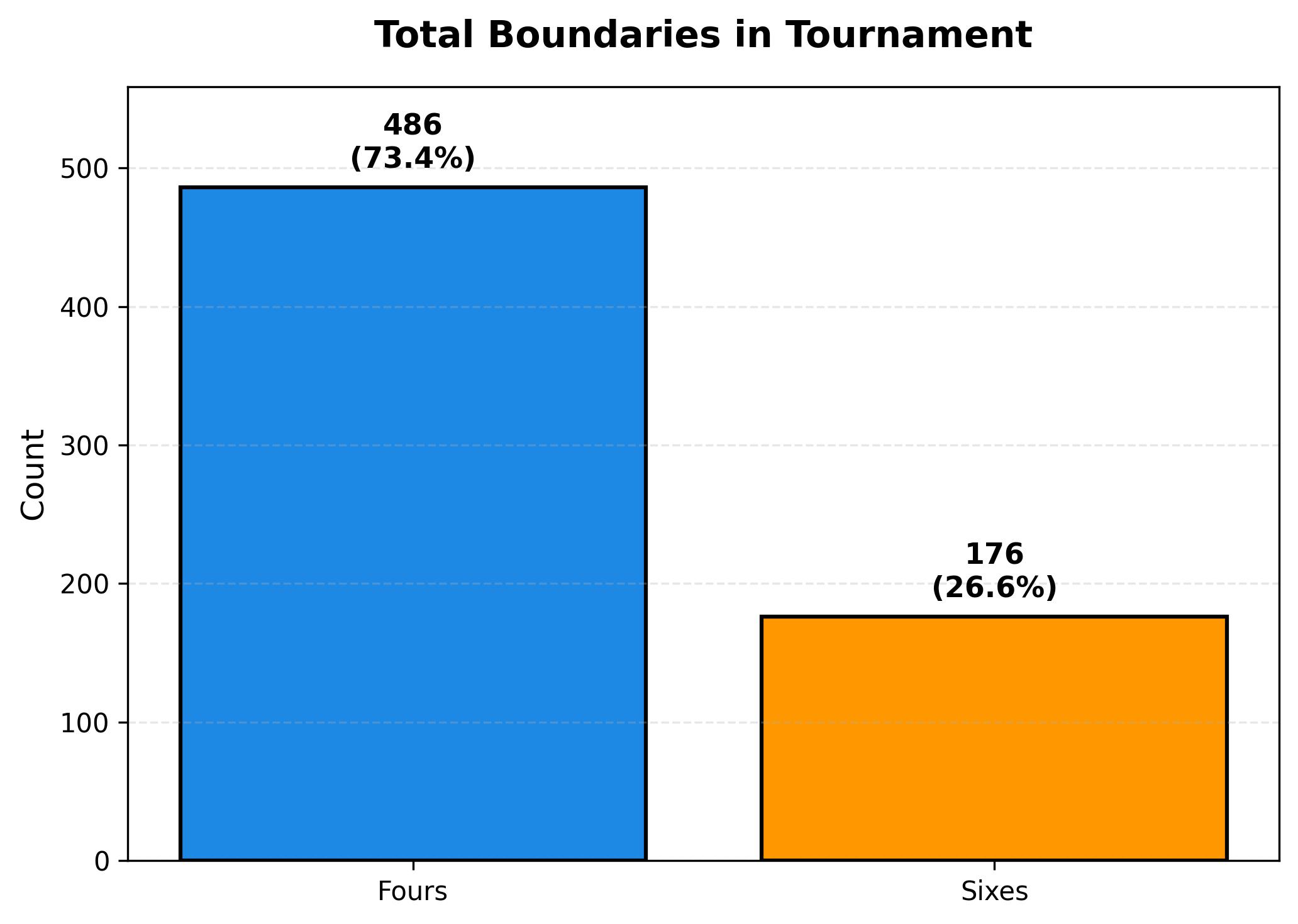}
\caption{Distribution of fours and sixes recorded during the tournament.}
\label{fig:boundaries}
\end{figure}


\section{Data Availability}
The dataset and accompanying analysis code are publicly available:
\begin{itemize}
\item \textbf{Zenodo Dataset:} \url{https://doi.org/10.5281/zenodo.17228056}
\item \textbf{GitHub Repository (EDA):} \url{https://github.com/kousarraza/AsiaCup2025}
\end{itemize}


\section*{Acknowledgments}
The author acknowledges \textbf{ESPNcricinfo} and the \textbf{Asian Cricket Council (ACC)} as the primary publicly available data sources used in compiling this dataset.


\end{document}